\documentclass[twoside]{article}
\usepackage[accepted]{icml2013}
\usepackage{natbib}

\usepackage{amsmath,amssymb,natbib,graphicx,color}

\newcommand{\R}{\mathbb{R}}
\newcommand{\Z}{\mathbf{Z}}
\newcommand{\Y}{\mathbf{Y}}
\newcommand{\U}{\mathbf{U}}
\newcommand{\UI}{u}
\newcommand{\X}{\mathbf{X}}
\newcommand{\XI}{x}
\newcommand{\Obs}{\mathbf{O}}
\newcommand{\OI}{o}
\newcommand{\Xib}{\boldsymbol{\xi}}

\begin{document}

\twocolumn[
\icmltitle{Group-sparse Embeddings in Collective Matrix Factorization}

\icmlauthor{Arto Klami}{arto.klami@cs.helsinki.fi}
\icmladdress{Helsinki Institute for Information Technology HIIT,
Department of Information and Computer Science, University of Helsinki}
\icmlauthor{Guillaume Bouchard}{guillaume.bouchard@xrce.xerox.com}
\icmladdress{Xerox Research Centre Europe}
\icmlauthor{Abhishek Tripathi}{abishek.tripathi3@xerox.com}
\icmladdress{Xerox Research Centre India}
            ]

\icmlkeywords{Collective matrix factorization, Variational Bayes, Bayesian learning, group-sparsity}

\begin{abstract}
 CMF is a technique for simultaneously learning low-rank representations 
 based on a collection of matrices with shared entities.
 A typical example is the joint modeling of user-item, item-property, and user-feature
  matrices in a recommender system. The key idea in CMF is that the
  embeddings are shared across the matrices, which enables transferring
  information between them. The existing solutions, however, break down
  when the individual matrices have low-rank structure not shared with others.
    In this work we present a novel CMF solution that allows
  each of the matrices to have a separate low-rank structure that is
  independent of the other matrices, as well as structures that 
  are shared only by a subset of them. 
  We compare MAP and variational Bayesian solutions based on 
  alternating optimization algorithms and show that the model
  automatically infers the nature of each factor using group-wise
  sparsity.
  Our approach supports in a principled way continuous, binary and count
  observations and is efficient for sparse matrices involving missing data. 
  We illustrate the solution on a number of examples, focusing in particular on an interesting use-case of
  augmented multi-view learning.
\end{abstract}

\section{INTRODUCTION}

\begin{figure}[t]
\centerline{\includegraphics[width=0.9\columnwidth]{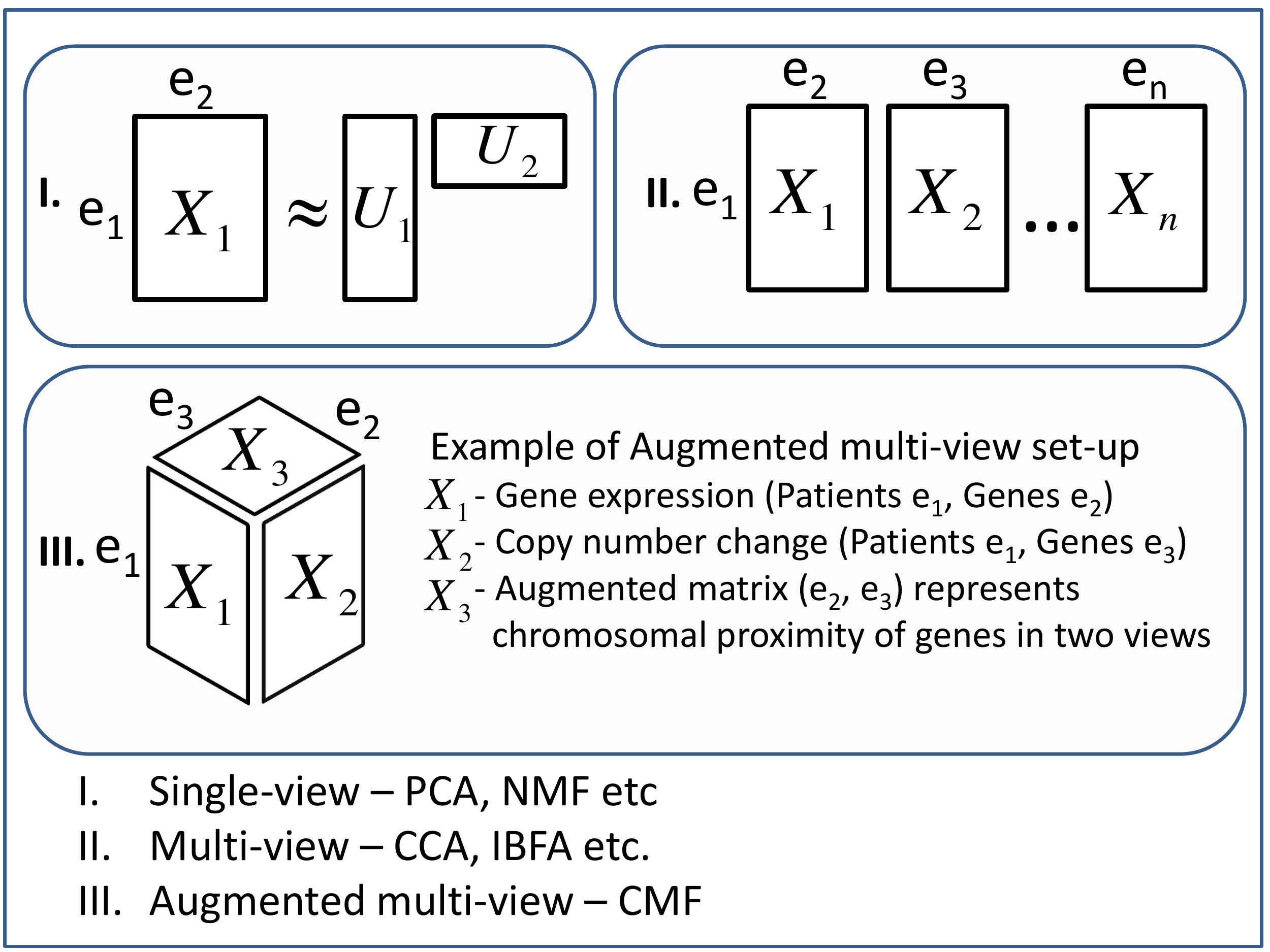}}
\caption{Examples of matrix factorization setups.}
\label{fig:examplesetups}
\end{figure}

Matrix factorization techniques provide low-rank vectorial representations
 by approximating a matrix $\X \in \R^{n \times d}$ as the
outer product of two rank-k matrices $\U_1 \in \R^{n \times k}$ and
$\U_2 \in \R^{d \times k}$ (Fig.~\ref{fig:examplesetups}-I). 
This formulation encompasses a multitude of standard data analysis models 
from PCA and factor analysis to more recent models such as NMF \citep{paatero1994positive,seung2001algorithms} 
and various sophisticated factorization models proposed for recommender 
system applications \citep{mnih2007probabilistic,koren2009matrix,sarwar2000application}.

Many data analysis tasks call for more complex setups.  
Multi-view learning (Fig.~\ref{fig:examplesetups}-II) considers scenarios with multiple matrices $\X_m$ that share
the same row entities but differ in the column entities; for example,
$\X_1$ might contain ratings given for $d_1$ different movies by $n$
different users, whereas $\X_2$ represents the same $n$ users with
$d_2$ profile features. For such setups the appropriate approach is to
factorize the set of matrices $\{\X_m\}$ simultaneously so that (at
least some of) the factors in $\U_1$ are shared across the matrices.
Models that share all of the factors are fundamentally equivalent to
simple factorizations of a concatenated matrix $\X=[\X_1,...,\X_m]$. To
reach a richer class of models one needs to allow each matrix to have
also private factors, i.e. factors independent of the other matrices
\citep{Jia10,Virtanen12aistats}. For the case of $M=2$ the distinction
is crystallized by the inter-battery factor analysis (IBFA) formulation of
\citet{Klami13jmlr}.

Even more general setups with arbitrary collections of matrices that
share some sets of entities have been proposed several times by different 
authors, under names such as co-factorization or multi-relational matrix factorization, 
and most end up being either a variant of tensor factorization of knowledge bases~\cite{nickel_three-way_2011, chen2013_learning} or 
a special case of \emph{Collective Matrix Factorization} \citep[CMF;][]{Singh08}. 
In this paper, we concentrate on the CMF model, i.e. on bilinear forms,
but the ideas can be easily extended to three-way interactions, i.e. tensors.
A prototypical example of CMF, illustrated by
\citet{Bouchard13aistats}, would be a recommender system setup where the target matrix
$X_1$ is complemented with two other matrices associating the users
and items with their own features. If the users and items are
described with the same features, for example by proximities to
geographical locations, the setup becomes circular. Another interesting use
case for such circular setups is found in augmenting multi-view learning,
in scenarios where additional information is provided on relationships
between the features of two (or more) views. Figure~\ref{fig:examplesetups}-III
depicts an example where the two views $\X_1$ and $\X_2$
represent expression and copy number alteration of the same patients.
Classical multi-view solutions to this problem would ignore the fact that
the column features for both views correspond to genes. With CMF, however, we can
encode this information as a third matrix $\X_3$ that provides
chromosomal promixity of the probes used for measuring the two views. Even
though this kind of setup is very common in practical multi-view learning,
the problem of handling such relationships has not attracted much attention.

Several solutions for the CMF problem have been presented.
\citet{Singh08} provided a maximum likelihood solution, \citet{Singh10} and
\citet{yin2013connecting}
used Gibbs sampling to approximate the posterior, and
\citet{Bouchard13aistats} presented a convex formulation of the
problem. While all of these earlier solutions to the CMF problem provide meaningful
factorizations, they share the same problem as the simplest solutions
to the multi-view setup; they assume that all of the matrices are
directly related to each other and that every factor describes
variation in all matrices. Such strong assumptions are unlikely
to hold in practical applications, and consequently the methods break
down for scenarios where the individual matrices have strong view-specific
noise or, more generally, any subset of the matrices has structure
independent of the others. In this work we remove the shortcoming by
introducing a novel CMF solution that allows also factors private to
arbitrary subsets of the matrices, by adding a group-wise sparsity
constraint for the factors.

We use group-wise sparse regularization of factors, where the groups 
corresponds to all the entities with the same type.
In the Bayesian setting, this group-regularization is obtained 
by using automatic relevance determination (ARD) for controlling
factor actity \citep{Virtanen12aistats}. 
This regularization enables us
to automatically learn the nature of each factor, resulting
in a solution free of tuning parameters. The model supports
arbitrary schemas for the collection of matrices, as well as multiple
likelihood potentials for various types of data (binary, count and
continous), using the quadratic lower bounds provided by
\citet{Seeger12} for non-Gaussian likelihoods.

To illustrate the flexibility of the CMF setup we discuss interesting
modeling tasks in Section~\ref{sec:usecases}. We pay
particular attention to the augmented multi-view learning setup
of Figure~\ref{fig:examplesetups}-III, showing that CMF provides
a natural way to improve on standard multi-view learning
when the different views lay in related
observation spaces. We also show experimentally the key advantage
of ARD used for complexity control, compared to 
computationally intensive cross-validation of regularization parameters.

\section{COLLECTIVE MATRIX FACTORIZATION}

Given a set of $M$ matrices $\X_m = [\XI_{ij}^{(m)}]$ describing relationships between $E$
sets of entities (with cardinalities $d_e$), the goal of CMF
is to jointly approximate the matrices with low-rank
factorizations. We denote by $r_m$ and $c_m$ the entity sets corresponding to the
rows and columns, respectively, of the $m$-th matrix.  For a simple matrix factorization we have $M=1$,
$E=2$, $r_m=1$, and $c_m=2$ (Fig.~\ref{fig:examplesetups}-I).  Multi-view setups, in turn, have
$E=M+1$, $r_m=1$ $\forall m$, and $c_m \in \{2,...,M+1\}$ (Fig.~\ref{fig:examplesetups}-II). 
Some non-trivial CMF setups are depicted in Figures~\ref{fig:examplesetups}-III and
\ref{fig:example-matrix-view}.

\subsection{Model}

We approximate each matrix with a rank-$K$ product plus additional
row and column bias terms. For linear models, the element corresponding to the row
$i$ and column $j$ of the $m$-th matrix is given by:
\begin{align}
\XI_{ij}^{(m)} = \sum_{k=1}^K \UI_{ik}^{(r_m)} \UI_{jk}^{(c_m)} + b_i^{(m,r)} + b_j^{(m,c)}
+ \varepsilon^{(m)}_{ij} \;,
\label{eq:CMF}
\end{align}
where $\U_e=[\UI_{ik}^{(e)}] \in \R^{d_e \times K}$ is the
low-rank matrix related to the entity set $e$, $b_i^{(m,r)}$ and $b_j^{(m,c)}$
are the bias terms for the $m$th matrix, and
$\varepsilon^{(m)}_{ij}$ is element-wise independent noise. We immediately see
that any two matrices sharing the same entity set use the same
low-rank matrix as part of their approximation, which enables sharing
information.

The same model can also be expressed in a simpler form by
crafting a single large symmetric observation matrix $\Y$ that contains all
$\X_m$, following the representation introduced by
\citet{Bouchard13aistats}. We will use this representation because
it allows implementing the private factors via group-wise sparsity.
We create one large entity set with $d =
\sum_{e=1}^E d_e$ entities and then arrange the observed matrices
$\X_m$ into $\Y$ such that the blocks not corresponding to any
$\X_m$ are left unobserved. The resulting $\Y$ is of size $d \times d$
but has only (at most) $\sum_{m=1}^M d_{r_m}d_{c_m}$ unique observed elements. In
particular, the blocks relating the entities of one type to themselves are not
observed.

\begin{figure}[t]
\centerline{\includegraphics[trim=1cm 0.7cm 2cm 0.5cm, width=0.8\columnwidth]{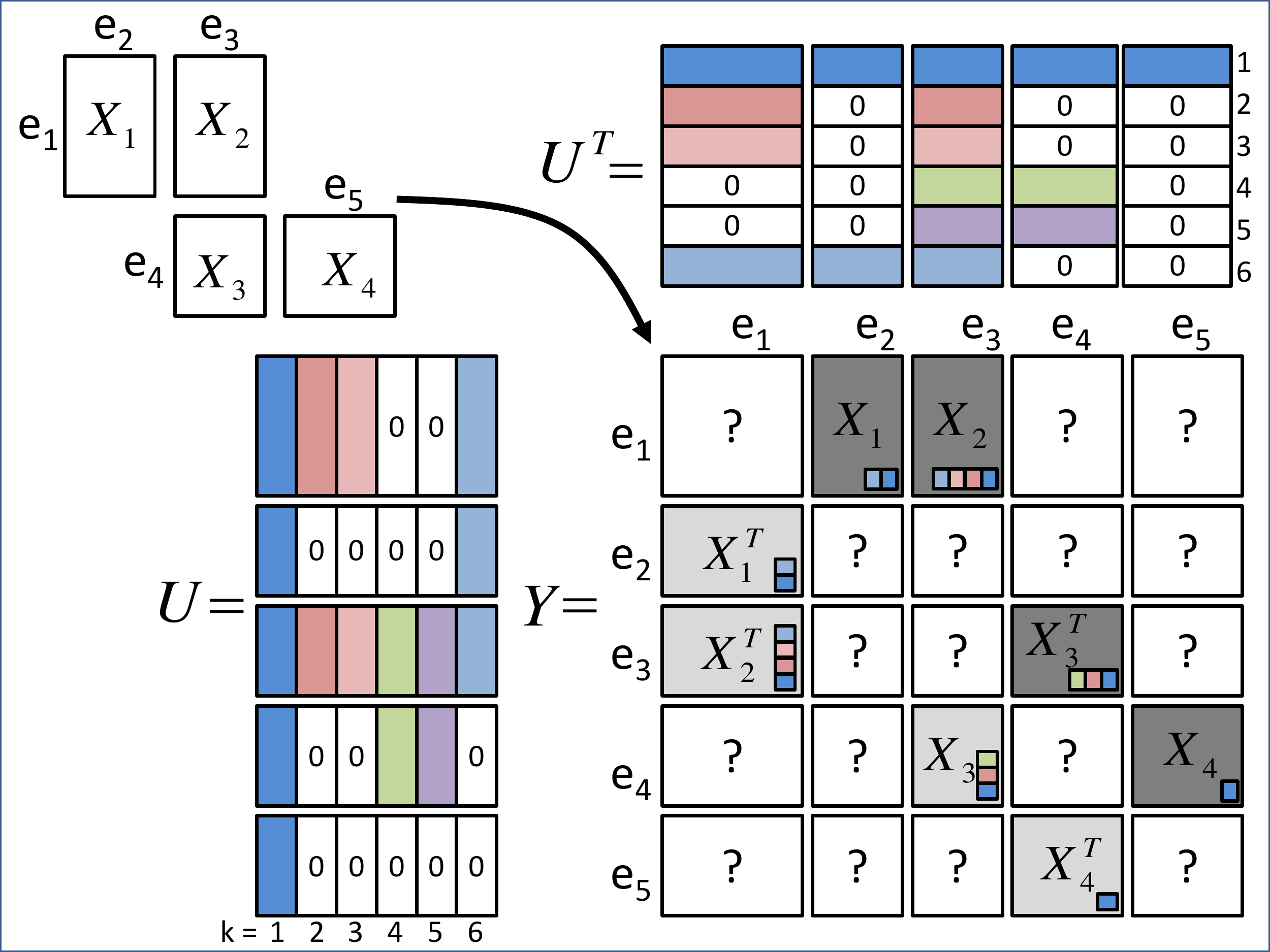}}
\caption{CMF setup encoded as a symmetric matrix factorization, with factors identified by colors.
The zero patterns in the $\U$ matrix induce private factors in the resulting $\Y$ matrix.
Contribution of factors are identified by small color patches next to the $\X$ matrices,
and the question marks (?) represent missing data.}
\label{fig:example-matrix-view}
\end{figure}

The CMF model can then be formulated as a symmetric matrix factorization
(see Figure~\ref{fig:example-matrix-view})
\begin{align}
\Y = \U \U^T + \boldsymbol{\varepsilon}, \label{eq:symmetric}
\end{align}
where $\U \in \R^{d \times K}$ is a column-wise concatenation of all of the different $\U_{e}$
matrices,
and the bias terms are dropped for notational simplicity. 
The noise $\boldsymbol{\varepsilon}$ is now symmetric but still independent 
over the upper-diagonal elements,
and the variance depends on the block the element belongs to.
Given this
re-formulation, any symmetric matrix factorization technique capable
of handling missing data can be used to solve the CMF problem; the
fact that the blocks along the diagonal are unobserved will usually be
crucial here, since it means that no quadratic terms will be involved
in the optimization. In Section~\ref{sec:inference}
a variational Bayesian approximation is introduced to learn the model, 
but before we explain how the basic formulation needs to be extended to
allow matrix-specific low-rank variations.

\section{Group-wise sparse CMF}

\subsection{Private factors in CMF}

Without further restrictions the solutions to \eqref{eq:symmetric} tie
all matrices to each other; for each factor $k$ the corresponding column of $\U$
has non-zero values for entities in every set $e$. This is
undesirable for many practical CMF applications where the individual
matrices are likely to have structured noise independent of other
matrices. Since the structured noise cannot be captured by the element-wise
independent noise terms $\boldsymbol{\varepsilon}$, the model will need to introduce
new factors for modeling the variation specific to one matrix alone.

We use the following property of the basic CMF model: if the $k$-th columns of
the factor matrices $\U_e$ are null for all but two entity types $r_m$ and $c_m$,
it implies that the $k$-th factor impacts only the matrix $\X_m$, i.e. the factor $k$
is a \emph{private} factor for relation $m$. To allow the automatic creation of these
private factors, we put group-sparse priors on the columns of the matrices $\U_e$. 
Using the symmetric representation, this approach creates group-sparse 
factorial representations similar to the one represented in Figure~\ref{fig:example-matrix-view}.
Note that if more than two groups of variables are non-zero for a given factor $k$, it
means that it is private for a group of matrices rather than a single matrix, and the standard CMF
is obtained if no groups equal to zero. In Figure~\ref{fig:example-matrix-view}
the first factor is a global factor
as used in the standard CMF, since it is non-zero everywhere, and the
rest are private to some matrices. Note that the last factor represented in light-blue in
($k=6$) is interesting because it is a private factor overlapping multiple matrices ($\X_1$ and $\X_2$)
rather than a single one for the other private factors (matrix $\X_1$ for factors 2 and 3, matrix $\X_3$ for factors 4 and 5).

To emphasize the group-wise sparsity structure in implementing
the private factors, we use the abbreviation gCMF for group-wise sparse CMF
i.e. a CMF model with this ability to learn separate private factors.

\subsection{Probabilistic model for gCMF}

We instantiate the general model by specifying Gaussian likelihood and
normal-gamma priors for the projections, so that in \eqref{eq:CMF} we
have
\begin{align*}
\varepsilon_{ij}^{(m)} &\sim {\cal N}(0,\tau_m^{-1}), 
&\tau_m &\sim {\cal G}(p_0,q_0), \\
\UI_{ik}^{(e)} &\sim {\cal N}(0,\alpha_{ek}^{-1}), 
&\alpha_{ek} &\sim {\cal G}(a_0,b_0).
\end{align*}
where $e$ is the entity set that contains the entity $i$. 
The crucial element here is the prior for $\U$.
Its purpose is to automatically select for each
factor a set of matrices for which it is active, which it does by
learning large precision $\alpha_{ek}$ for factors $k$ that
are not needed for modeling variation for entity set $e$.
In particular, the prior takes care of matrix-specific low-rank structure,
by learning factors for which $\alpha_{ek}$ is small for only two
entity sets corresponding to one particular matrix.

For the bias terms we use a hierarchical prior
\begin{align*}
b_i^{(m,r)} &\sim {\cal N}(\mu_{rm},\sigma_{rm}^2), 
&b_j^{(m,c)} &\sim {\cal N}(\mu_{cm},\sigma_{cm}^2), \\
\mu_{\cdot m} &\sim {\cal N}(0,1), &\sigma_{\cdot m}^2 &\sim {\cal U}[0,\infty].
\end{align*}
The hierarchy helps especially in
modeling rows (and equivalently columns) with lots of missing data,
and in particular provides reasonable values also for rows with no
observations (the cold-start problem of new users in recommender
systems) through $\mu_{rm}$.

\section{LEARNING}
\subsection{MAP solution}
Providing a MAP estimate for the model is straighforward,
but results in a practical challenge of needing to choose the hyper-parameters
$\{a_0,b_0,p_0,q_0\}$, usually through cross-validation.
This is particularly difficult for setups with several heterogeneous
data matrices on arbitrary scales. Then large hyper-priors are needed
for preventing overfitting, which in turn makes it difficult to push
$\alpha_{ek}$ to sufficiently large values to make the factors private
to subsets of the matrices. Hence, we proceed to explain more reasonable
variational approximation that avoids these problems.

\subsection{Variational Bayesian inference}
\label{sec:inference}

It has been noticed that Bayesian approaches which take into account 
the uncertainty about the values of the latent variables lead to increased 
predictive performance~\cite{Singh10}. Another important advantage of 
Bayesian learning is the ability to automatically select regularization parameters by maximizing the data evidence. 
While existing Bayesian approaches for CMF used MCMC techniques 
for learning, we propose here to use variational Bayesian learning (VB)
by minimizing the KL divergence between a tractable approximation
and the true observation probability.
We use a fully factorized approximation similar to 
what \citet{Ilin10} presented for Bayesian PCA with missing data, and
implement non-Gaussian likelihoods using the quadratic bounds
by \citet{Seeger12}. In the following we will summarize the
main elements of the algorithm, leaving some of the technical details
to these original sources.

\paragraph{Gaussian observations}

For Gaussian data we approximate the posterior with
\begin{align}
Q(\Theta) &=
 \left [
 \prod_{e=1}^E 
\prod_{k=1}^K \left (
q(\alpha_{ek})
\prod_{i=1}^{d_e} q(\UI_{ik}^{(e)})
\right )
\right ] \label{eq:cost}\\
& 
\hspace{-0.7cm}
\left [
\prod_{m=1}^M 
q(\tau_m) q(\mu_{rm}) q(\mu_{cm})
\prod_{i=1}^{d_{r_m}} q(b_i^{(m,r)})
\prod_{j=1}^{d_{c_m}} q(b_j^{(m,c)})
\right ]. \notag
\end{align}
Here $q(\alpha)$ and $q(\tau)$ are Gamma distributions, whereas the
others are normal distributions. For all other parameters we use
closed-form updates, but $\bar \U_{e}$, the mean parameters of
$q(\U_{e})$, are updated with Newton's method for each
factor at a time. The gradient-based updates are used because for
observation matrices with missing entries closed-form
updates would be available only for each element $\bar \UI_{ik}^{(e)}$
separately, which would result in very slow convergence
\citep{Ilin10}. The update rules for $Q(\Theta)$ are in the supplementary
material.

\paragraph{Non-Gaussian observations}
For non-Gaussian data we use the approximation schema presented by
\citet{Seeger12}, adaptively approximating non-Gaussian
likelihoods with spherical-variance Gaussians. This allows an
optimization scheme that alternates between two steps: (i) updating
$Q(\Theta)$ given pseudo-data $\Z$ (which is assumed Gaussian), and (ii)
updating the pseudo-data $\Z$ by optimizing a quadratic term
lower-bounding the desired likelihood potential. The full derivation
of the approach is provided by \citet{Seeger12}, but the resulting
equations as applied to gCMF are summarized below. We update the
pseudodata with
\begin{align*}
\Xib_m &= E[\U_{r_m}] E[\U_{c_m}]^T, \\
\Z_m &= (\Xib_m - f_m'(\Xib_m)/\kappa_m),
\end{align*}
where the updates are element-wise and independent for each matrix.
Here $f_m'(\Xib_m)$ is the derivative of the $m$-th link function $-\log p(\X_m|\U_{r_m}\U_{c_m}^T)$
and $\kappa_m$ is the maximum value of the second derivative of the same
function. Given the pseudo-data $\Z$, the approximation $Q(\Theta)$ can
be updated as in the Gaussian case, using $\tau_m=\kappa_m$ as the
precision. Note that the link functions can be different for different
observation matrices, which adds support for heterogeneous data;
in Section~\ref{sec:experiments} we illustrate binary and count data.

\section{RELATED WORK}

For $M=1$ the model is equivalent to Bayesian (exponential family) PCA. In particular, it
reduces to gradient-based optimization for the model by
\citet{Seeger12}. For this special case it is typically advisable to
use their SVD-based algorithm,
since it provides closed-form solution for the Gaussian case.

For multi-view setups where every matrix shares the same row-entities
the model equals Bayesian inter-battery factor analysis
(when $M=2$) \citep{Klami13jmlr} and its extension group-factor
analysis (when $M>2$) \citep{Virtanen12aistats}. However, our
inference solution has a number of advantages.
In particular, our solution supports
wider range of likelihood potentials and provides efficient
inference for missing data. These improvements suggests that the
proposed algorithm should be preferred over the earlier solutions.

The most closely related methods are the earlier CMF
solutions, in particular the ones presented in the probabilistic
framework. The early solutions by \citet{Lippert08} and \citet{Singh08}
provide only maximum-likelihood solutions, whereas 
\citet{Singh10} provided fully Bayesian solution by formulating
CMF as a hierarchical model. They use normal-Inverse-Wishart priors
for the factors, with spherical hyper-prior for the Inverse-Wishart
distribution. This implies each factor is assumed to be roughly
equally important in describing each of the matrices, and that their
model will not provide matrix-specific factors as our model does.
For inference they use computationally heavy Metropolis-Hastings.
Their model also supports arbitrary likelihood potentials
and arbitrary CMF schemas, though their experiments are limited
to cases with $M=2$.

\section{USE CASES}
\label{sec:usecases}

Even though CMF is widely applicable to factorization of arbitrary
matrix collections, it is worth describing some typical setups
to illustrate common use cases where data analysis practitioners
might find it useful.

\paragraph{Augmenting multi-view learning}
In multi-view learning (Fig.~\ref{fig:examplesetups}-II) the row
entities are shared, but the column entities in different views
are arbitrary. In
many practical applications, however, the column entities share some obvious
relationships that are ignored by the multi-view matrix factorization
models. A common example considers computing CCA between two
different high-throughput systems biology measurements of the same
patients, so that both matrices are patients times genes
\citep[see, e.g.,][]{Witten09}. In natural language processing, in turn, we
have setups with different languages as row entities and words as
column entities \citep{tripathi2010bilingual}. In both cases there are obvious
relationships between the column features. In the first example it
is an identity relation, whereas in the latter lexigographic or
dictionary-based information provides proximity relations for the
column entities.  Yet another example can be imagined in joint
analysis of multiple brain imaging modalities; the column entities
correspond to brain regions that have spatial relationships even
though the level of representation might be very different when, e.g.,
analyzing fMRI and EEG data jointly \citep{Correa10}.

Such relationships between the column entities can easily be taken into
account with CMF using the cyclical relational schema of
Figure~\ref{fig:examplesetups}-III. We call this approach
\emph{augmented multi-view learning}.
We can encode any kind of similarity between the features
as long as the resulting matrix can reasonably be modeled as low-rank.
In the experimental section we will demonstrate setups where the
features live in a continuous space (genes along the chromosome, pixels
in a two-dimensional space) and hence we can measure distances between
them. We then convert these distances into binary promixity relationships,
to illustrate that already that is sufficient for augmenting
the learning.

\paragraph{Recommender systems}

The simplest recommender systems seek to predict missing entries in a
matrix of ratings or binary relevance indicators~\citep{koren2009matrix}.  
The extensive
literature on recommender systems indicates that incorporating
additional information on the entities helps making such predictions
\citep{stern2009matchbox,fang2011matrix}. CMF is a natural way of encoding such information, in form of additional matrices between the entities of interest and
some features describing them.

\begin{figure*}[t]
\begin{center}
\begin{tabular}{cc}
\includegraphics[width=0.7\columnwidth]{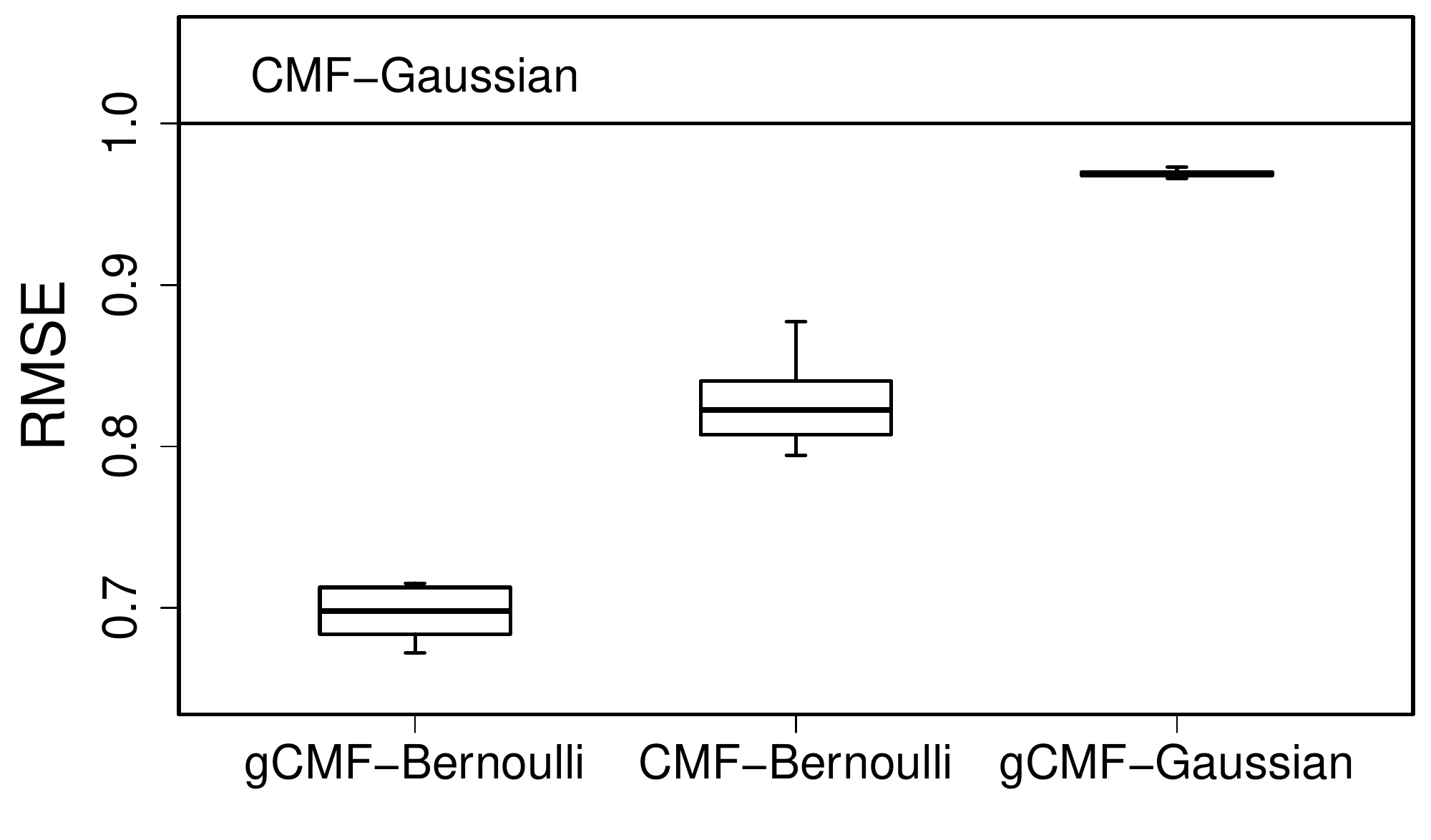} &
\includegraphics[width=0.7\columnwidth]{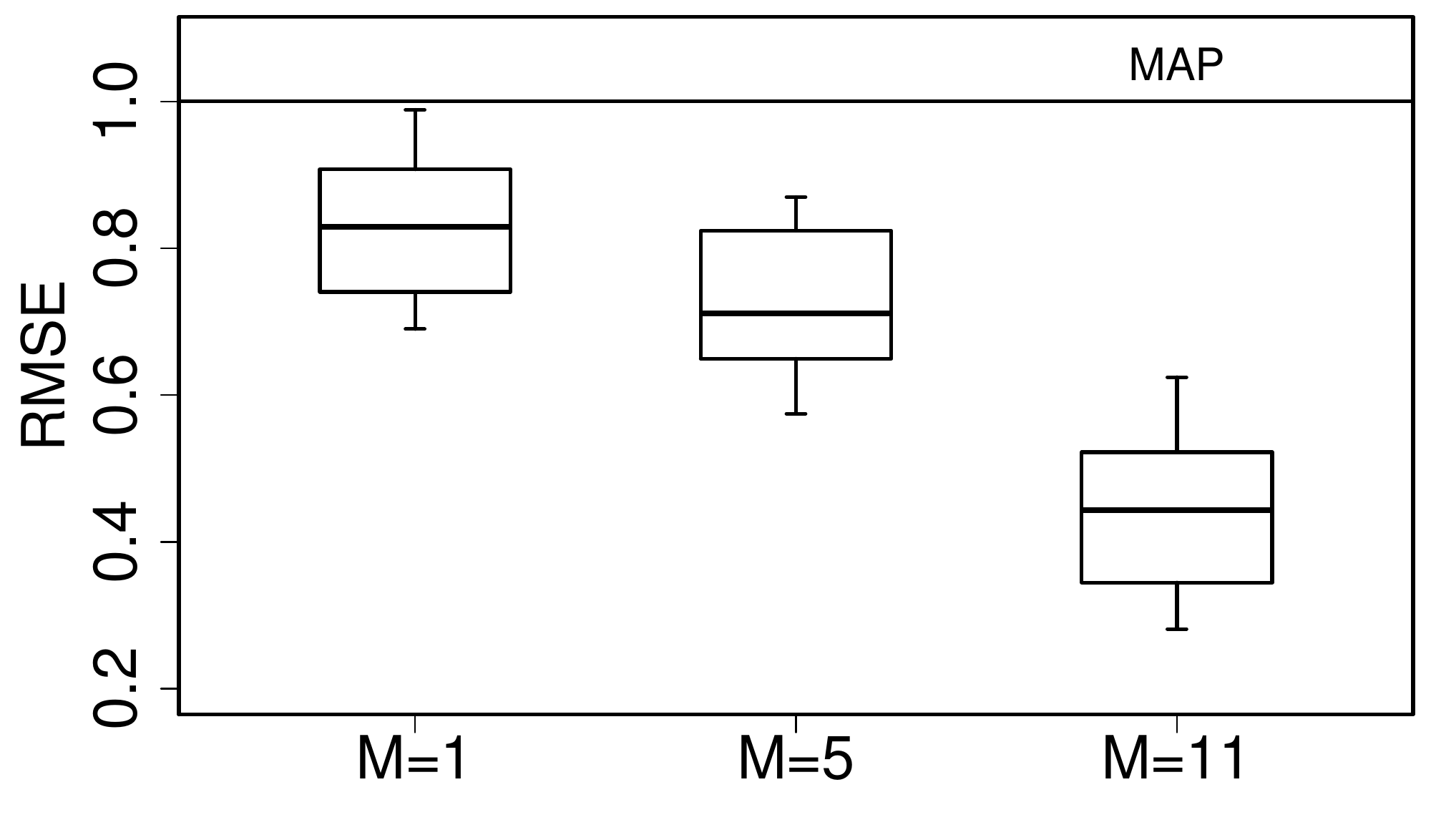}
\end{tabular}
\end{center}
\caption{{\bf Left:} Relative error for a circular setup
of $M=5$ binary matrices (see text for details), scaled so that
CMF with Gaussian likelihood has error of one. The correct
likelihood helps for both gCMF and CMF and modeling the private
factors helps for both likelihoods, the combined gain of both
aspects being $30\%$. The results are similar for other values of $M>1$.
{\bf Right:} Relative error of VB vs MAP, scaled so that
zero corresponds to the ground truth and one to the error of the
MAP solution. For small $M$ MAP can still compete (though it is
worse than VB already for $M=1$), but for large $M$ it becomes
worthless; for $M=11$ VB reduces the error to roughly
half. Furthermore, VB requires no tuning parameters, whereas for
the MAP solution we needed to perform cross-validation over two
regularization parameters.}
\label{fig:toy1}
\end{figure*}

While many other techniques can also be used for incorporating
additional information about the entities, the CMF formulation opens
up two additional types of extra information not easily implemented by
the alternative means. The first is a circular setup where both the
row and column entities of the matrix of interest are described by the
same features \citep{Bouchard13aistats}. This is typically the case for example in social
interaction recommenders where both rows and columns correspond to human
individuals. 
The
other interesting formulation uses higher-order auxiliary data. For
example, the movies in a classical recommender system can be
represented by presence of actors, whereas the actors themselves are
then represented by some set of features. This leads to a chain of
matrices providing more indirect information on the relationships
between the entities.

\section{EXPERIMENTS}
\label{sec:experiments}

We start with technical validations showing
the importance of choosing the correct likelihood potential and incorporating
private factors in the model, as well as the advantages variational approximation
provides over MAP estimation. We then proceed to show how CMF outperforms
classical multi-view learning methods in scenarios where we can augment
the setup with between-feature relationships.

Since the main goal is to demonstrate the conceptual importance of
solving the CMF task with private factors, we use special cases
of gCMF as comparison methods. This helps to show that
the difference is really due to the underlying idea instead of the
inference procedure; for example, when comparing against
\citet{Singh10} the effects could be masked by
differences between Metropolis-Hastings and variational approximation
that are here of secondary importance.

The closest comparison method, denoted by CMF, is obtained by forcing $\alpha_{ek}$
to be a constant $\alpha_k$ for every entity type $e$. It corresponds to the VB solution
of the earlier CMF models and hence does not support private factors.
For the augmented multi-view setup we will also compare against the
special cases of gCMF and CMF that use only two matrices over the three
entity sets, denoting them by CCA and PCA, respectively. Finally,
in one experiment we will also compare against gCMF without the bias
terms, to illustrate their importance in recommender systems. For all
methods we use sufficiently large $K$, letting ARD prune out unnecessary components,
and run the algorithms until the variational lower bound converges.
We measure the error by root mean square error (RMSE), relative to
one of the methods in each experiment.

\subsection{Technical illustration}

We start by demonstrating the difference between the proposed model
and classical CMF approaches on an artificial data. We sample $M$
binary matrices that form a cycle over $M$ entity sets (of sizes $100-150$),
so that the first matrix is between the entity sets 1 and 2,
the second between the entity sets 2 and 3, and finally the last one
is between the $M$-th and first entity set. We generate datasets that have
$5$ factors shared by all matrices plus two factors of low-rank noise
specific to each matrix. This results in $5+2M$ true factors, and we
learn the models with $10+2M$ factors, letting ARD prune out
the extra ones.

Figure~\ref{fig:toy1} (left) shows the accuracy in predicting
the missing entries ($40\%$ of all) for gCMF as well as
a standard CMF model. For both models we show the results for both
(incorrect) Gaussian and Bernoulli likelihoods. The experiment
verifies the expected results: Using the correct likelihood improves
the accuracy, as does correctly modeling private noise factors.

We use the same setup to illustrate the importance of using
variational approximation for inference, this time with Gaussian
noise and entity set sizes between $40-80$.  For MAP we
validate the strength of the Gamma hyper-priors for $\tau$ and
$\alpha$ over a grid of $11 \times 11$ values for $a_0=b_0$ and
$p_0=q_0$, using two-fold
cross-validation within the observed data. In total we hence need to
run the MAP variant more than 200 times to get the result, in contrast
to the single run of the VB algorithm with vague
priors using $10^{-10}$ for every parameter. Figure~\ref{fig:toy1} (right) shows that despite heavy
cross-validation the MAP setup is always worse and the gap gets bigger
for more complex setups. This illustrates how the VB solution with no
tunable hyperparameters is even more crucial for CMF than it would be
for simpler matrix factorizations. For MAP using the same
hyper-priors for all matrices necessarily becomes a compromise
for matrices of different scales, whereas validating separate scales
for each matrix would be completely infeasible (requiring validation
over $2M$ parameters).

\subsection{Augmented multi-view learning}

We start with a multi-view setup in computational biology, using data
from \citet{Pollack02} and the setup studied by \citet{Klami13jmlr}.  The
samples are 40 patients with breast cancer, and the two views correspond
to high-throughput measurements of expression and copy number alteration
for 4287 genes. We compare the models in the task of predicting random
missing entries in both views, as a function of the proportion of
missing data.

\begin{figure}[t]
\centerline{\includegraphics[width=0.75\columnwidth]{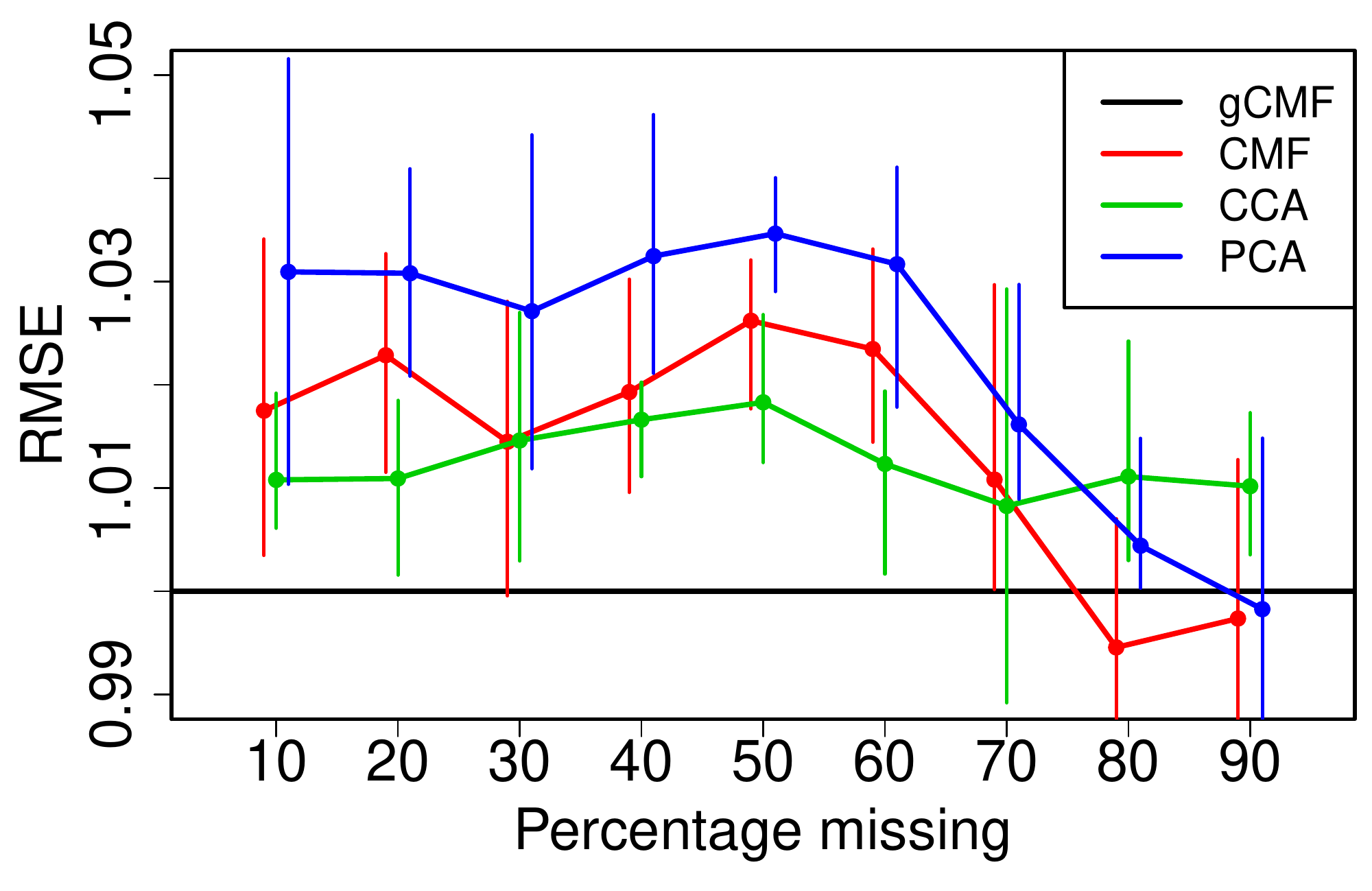}}
\caption{
Relative prediction error for augmented multi-view gene experiment,
scaled so that gCMF has error one and is represented by the
horizontal black line.
For reasonable amounts of missing data (x-axis) the methods with
private factors (gCMF and CCA) outperform the ones without, and
modeling the
proximity relationship between the genes (gCMF and CMF)
improves the accuracy. The confidence intervals correspond to
10\% and 90\% quantiles over random choices of missing data.}
\label{fig:gene}
\end{figure}

The multi-view methods use the data as such, whereas the CMF variants
also use a third $d_2 \times d_3$ matrix that encodes the proximity
of the genes in the two views. It is a binary matrix such that
$\XI_{i,j}^{(3)}$ is one with probability $\exp(-|l_i-l_j|)$, where 
$l_i$ is
the chromosomal location measured in $10^7$ basepairs. This
encodes the reasonable assumption that
copy number alterations are more likely to influence the expression
of nearby genes. Figure~\ref{fig:gene} shows how this information
helps in making the predictions. For reasonable amounts of missing
data, gCMF is consistently the best method, outperforming both CMF
as well as the standard multi-view methods. For extreme
cases with at least $80\%$ missing data the advantage is finally lost. The
importance of the private factors is seen also in CCA outperforming
PCA, whereas CMF and CCA are roughly as accurate; both include one
of the strenghts of gCMF.

In another example we model images of faces taken in two alternative
lighting conditions, but from the same viewing angle. We observe the
raw grayscale pixels values of $50 \times 50$ images, and for the CMF
methods we use a third matrix (size $2500 \times 2500$, of which
random $10\%$ is observed) to encode
proximity of pixels in the two views, using Gaussian kernel to
provide the probability of one for a binary relation.  We train the
model so that we have observed 6 images in both views and then 7
images for each view alone, for a total of 20 images. The task is to
predict the missing views for these images, without any observations.

\begin{figure}[t]
\centerline{\includegraphics[width=0.75\columnwidth]{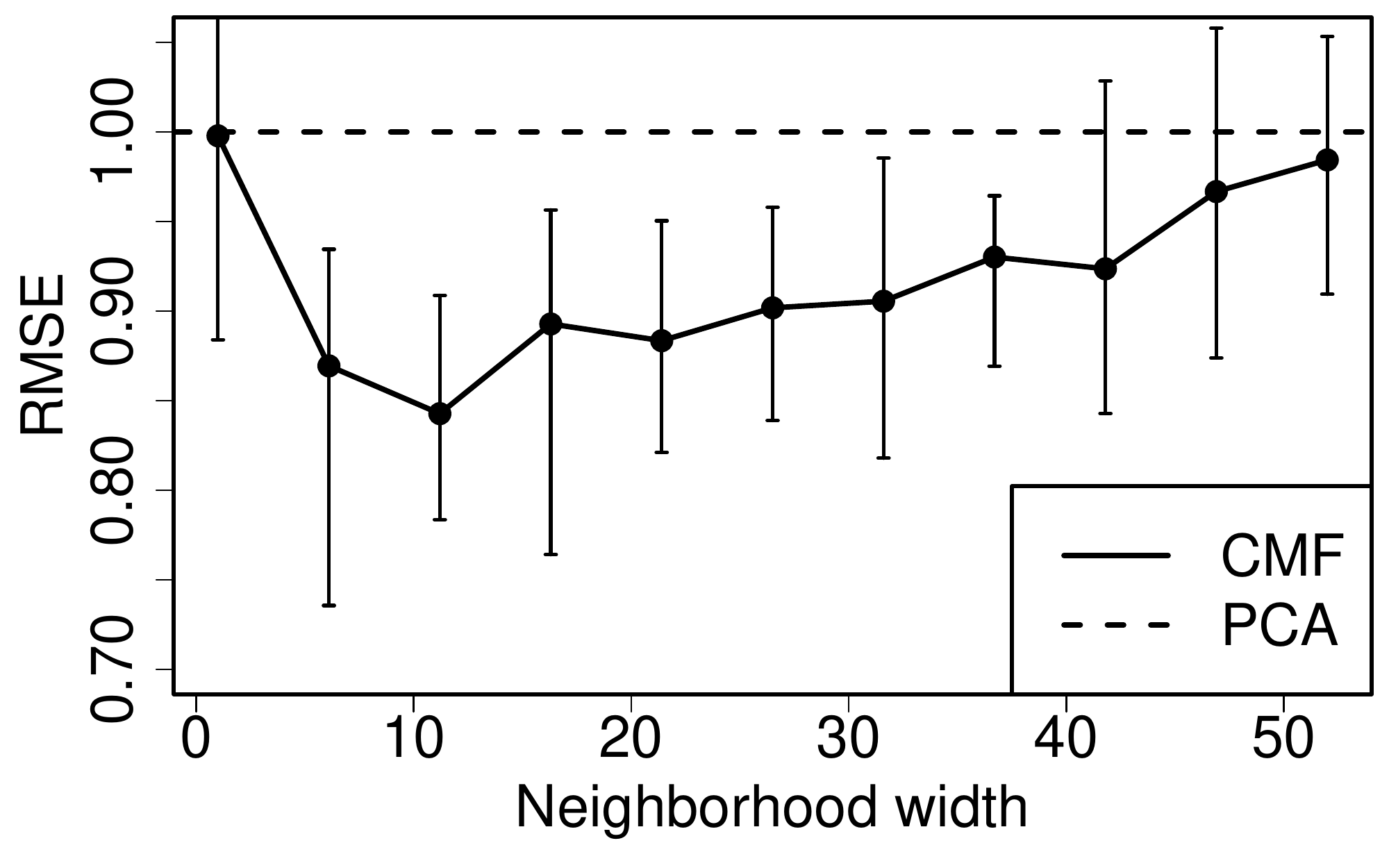}}
\caption{Prediction error for a multi-view image reconstruction task as a function of the
neighborhood width in constructing the proximity augmentation view. The augmentation helps for a wide range of
promixity relationships, and the solution reverts back to the non-augmented
accuracy for very narrow and wide neighborhoods.}
\label{fig:faces}
\end{figure}

Figure~\ref{fig:faces} plots the prediction errors as a function
of the neighborhood $\sigma$ used in constructing the promity
relationships. We see that for very narrow and very wide neighborhoods
the CMF approach reverts back to the classical multi-view model, since
the extra view consists almost completely of zeros or ones, respectively.
For proper neighborhood relationships the accuracy in predicting
the missing view is considerably improved.

\begin{table*}[t]
\caption{RMSE for two recommender system setups, with boldface indicating the best results. The results for
convex CMF (CCMF) are taken from \citet{Bouchard13aistats} for the best
regularization parameter values. Our model provides comparable result without
the bias terms, without needing any tuning for the parameters, and
the bias terms helps considerably with the cold-start problem
especially in MovieLens. Without bias terms gCMF also outperforms CMF for
all cases, but with the bias terms the methods are practically identical
for these data sets. This suggests these data sets do not have strong private
structure that could not be modeled with the bias terms alone. It is important
to note that allowing for the private factors never hurts; gCMF is always
at least as good as CMF.}
\label{tbl:recsys}
\begin{center}
\begin{tabular}{l|l|cccccc}
Data & MovieLens & \multicolumn{5}{c}{Flickr} \\
Relation & $\X_1$-Count & $\X_1$-Binary & $\X_2$-Binary & $\X_3$-Binary & $\X_4$-Binary & $\X_5$-Gaussian \\
\hline
CCMF (reg=10) & 1.0588 & 0.7071 & 0.2473 & 0.3661 & 0.2384 & {\bf 1.0033}\\
CMF without bias & 1.0569 & 0.5120 & 0.2324 & 0.5093 & 0.2176 & 1.0092\\
CMF with bias & 0.9475 & {\bf 0.5000} & 0.2369 & {\bf 0.2789} & {\bf 0.2109} & {\bf 1.0033}\\
gCMF without bias & 1.0418 & 0.5003 & {\bf 0.2291} & 0.5014 & 0.2167 & 1.0039\\
gCMF with bias & {\bf 0.9474} & {\bf 0.5000} & 0.2369 & {\bf 0.2789} & {\bf 0.2109} & {\bf 1.0033}\\
\end{tabular}
\end{center}
\end{table*}

\subsection{Recommender systems}

Next we consider classical recommender systems, using MovieLens
and Flickr data as used in earlier CMF experiments by
\citet{Bouchard13aistats}. We compare gCMF with the convex CMF
solution presented in that paper, showing that it finds the same
solution when the bias terms are turned off (Table~\ref{tbl:recsys}).
We also illustrate that modeling the bias terms explicitly is as
useful for CMF as it has been shown to be for other types of
recommender systems. To our knowledge gCMF is the first
CMF solution with such bias terms.

Both data sets have roughly 1 million observed entries, and
our solutions were computed in a few minutes on a laptop.
The total computation time is hence roughly comparable to the
times \citet{Bouchard13aistats} reported for CCMF using one
choice of regularization parameters. Full CCMF solution is
considerably slower since it has to validate over them.

\section{DISCUSSION}

Collective matrix factorization is a very general technique for
revealing low-rank representations for arbitrary matrix collections.
However, the practical applicability of earlier solutions has been
limited since they implicitly assume all factors
to be relevant for all matrices. Here we presented a general technique
for avoiding this problem, by learning the CMF solution as symmetric
factorization of a large square matrix while enforcing group-wise
sparse factors.

While any algorithm aiming at such sparsity structure will provide
shared and private factors for a CMF, the variational
Bayesian solution presented in this work has some notable advantages.
It is more straighforward than the sampling-based
alternative by \citet{Singh10} (which could be modified to
incorporate private factors) while being free of tunable
regularization parameters required by the convex solution of
\citet{Bouchard13aistats}. The model also subsumes some earlier models
and provides extensions for them. In particular, it can be used
to efficiently learn Bayesian CCA solution for missing data and
non-conjugate likelihoods, providing the first
efficient Bayesian CCA
between binary observations.

One drawback of CMF is its inability to handle multiple relations accross 
two entity type. Tensor factorization methods alleviate this problem,
as illustrated in the recent work on multi-relational data~\cite{glorot2013_semantic,chen2013_learning}.

\subsection*{Acknowledgments}

We acknowledge support from the University Affairs Committee of the
Xerox Foundation. AK was also supported by Academy
of Finland (grants 251170 and 266969) and Digile SHOK project D2I.

\bibliographystyle{plainnat}
\bibliography{iclr14}

\newpage

\twocolumn[
\begin{@twocolumnfalse}
\icmltitle{Group-sparse Embeddings in Collective Matrix Factorization\\
\vspace{0.5cm}
Supplementary material}
\end{@twocolumnfalse}
]

This supplementary material for the manuscript ``Group-sparse
Embeddings in Collective Matrix Factorization'' provides
more details on the variational approximation described in the
paper.

\section*{Notation}

The factors in \eqref{eq:cost} are
\begin{align*}
q(\UI_{ik}^{(e)}) &= {\cal N}(\bar \UI_{ik}^{(e)},\tilde \UI_{ik}^{(e)}), \\
q(\alpha_{ek}) &= {\cal G}(a_{ek},b_{ek}),  \;\;
q(b_i^{(m,r)}) = {\cal N}(\bar b_i^{(m,r)}, \tilde b_i^{(m,r)}), \\
q(\tau_{m}) &= {\cal G}(p_m,q_m), \;\;
q(b_j^{(m,c)}) = {\cal N}(\bar b_j^{(m,c)}, \tilde b_j^{(m,c)}).
\end{align*}
and we denote by $\bar \alpha$ and $\bar \tau$ the expectations of $\alpha$
and $\tau$. The observed entries in $\X_m$
are given by $\Obs_m \in [0,1]^{d_{r_m} \times d_{c_m}}$,
with $n_m = \sum_{ij} \OI_{ij}^{(m)}$ indicating their total number.
Finally, we denote $\hat \XI_{ij}^{(m)} = \left (
\XI_{ij}^{(m)} - \sum_{k=1}^K \bar \UI_{ik}^{(r_m)} \bar \UI_{jk}^{(c_m)} - \bar b_i^{(m,r)} - \bar b_j^{(m,c)}
\right )$.

\section*{Algorithm}

The full algorithm repeats the following steps until convergence.
\begin{enumerate}
\item For each entity set $e$, compute the gradient of $\bar \U_e$
  using \eqref{eq:Ugradient} and compute the variance parameter
  $\tilde \U_e$ using \eqref{eq:Uvariance}.
\item Update $\bar \U_e$ with under-relaxed Newton's step. The
  element-wise update is $\bar \UI_{ik}^{(e)} \leftarrow
  (1-\lambda) \bar \UI_{ik}^{(e)} + \lambda (\tilde \UI_{ik}^{(e)})^{-1} g_{ik}^{(e)}$
  with $0<\lambda<1$ as the regularization parameter.
\item Update the approximations for the bias terms using \eqref{eq:bias}.
\item Update the approximations for the automatic relevance determination parameters using
  \eqref{eq:alpha}.
\item For all matrices $\X_m$ with Gaussian likelihood, update the 
  approximations for the noise
  precision parameters using \eqref{eq:tau}. For all matrices $\X_m$ with
  non-Gaussian likelihood, update the pseudo-data using
  \eqref{eq:pseudodata}.
\end{enumerate}

\section*{Details}

\paragraph{Updates for the factors:}

The gradient with respect to the mean parameters of the factors
is computed as
\begin{align}
g_{ik}^{(e)} &= \bar \alpha_{ek} \bar \UI_{ik}^{e} + \notag \\
&\sum_{m;r_m=e}
\bar \tau_m \sum_j \left [
-\hat \XI_{ij}^{(m)} \bar \UI_{jk}^{(c_m)} + \bar \UI_{ik}^{(e)} \tilde \UI_{jk}^{(c_m)}
\right ] \label{eq:Ugradient} \\
&\sum_{m;c_m=e}
\bar \tau_m \sum_j \left [
-\hat \XI_{ij}^{(m)} \bar \UI_{ik}^{(r_m)} + \bar \UI_{jk}^{(e)} \tilde \UI_{ik}^{(r_m)}
\right ]. \notag
\end{align}

For $\tilde \U_e$ we have closed-form updates
\begin{align}
\tilde \UI_{ik}^{(e)} & = \left [
\bar \alpha_{ek} + 
\sum_{m;c_m=e} \bar \tau_m \sum_j \left ( 
(\bar \UI_{jk}^{(r_m)})^2 + \tilde \UI_{jk}^{(r_m)}
\right ) \right.\notag \\
&\left. + \sum_{m;r_m=e} \bar \tau_m \sum_j \left ( 
(\bar \UI_{jk}^{(c_m)})^2 + \tilde \UI_{jk}^{(c_m)}
\right )
\right ]^{-1}. \label{eq:Uvariance}
\end{align}

\paragraph{Updates for the bias terms:}

The approximations for the row bias terms are updated as
\begin{align}
\tilde b_i^{(m,r)} &= \left (\bar \tau_m \sum_j \OI_{ij}^{(m)} + \sigma^{-2} \right )^{-1}, \label{eq:bias} \\
\hat b_i^{(m,r)} &= \tilde b_i^{(m,r)} \left (
\bar \tau_m \mu_i + \mu_{rm}/\sigma_{rm}^2
\right ), \notag
\end{align}
where $\mu_i$ is a shorthand notation for the mean of $\XI_{ij}^{(m)}
- \sum_k \bar \UI_{ik}^{(m)} \bar \UI_{jk}^{(m)} - \bar b_j^{(m,c)}$
over the observed entries.  We additionally update $q(\mu_{rm})$ using
standard variational update for Gaussian likelihood and prior, and use
point estimate for $\sigma_{rm}^2$. The updates for the column bias
terms follow naturally.

\paragraph{Updates for the ARD terms:}

The approximations for the ARD variance parameter terms are updated as
\begin{align}
a_{ek} & = a_0^{\alpha} + d_e/2, \label{eq:alpha} \\
b_{ek} & = b_0 + 0.5 \sum_{i=1}^{d_s} \sum_{k=1}^K
\left ( (\bar \UI_{ik}^{(e)})^2 + \tilde \UI_{ik}^{(e)} \right ). \notag
\end{align}

\paragraph{Updates for the precision terms:}

For each matrix with Gaussian likelihood the approximation for
the precision term is updated as
\begin{align}
p_m &= p_0 + n_m/2, \label{eq:tau} \\
q_m &= q_0 + \frac{1}{2n_m} \sum_{ij} \bigg [ 
(\hat \XI_{ij}^{(m)})^2
 + \tilde b_i^{(m,r)} + \tilde b_j^{(m,c)} \notag \\
& \hspace{-1em}\left. + \sum_{k=1}^K
\left (
(\bar \UI_{ik}^{(r_m)})^2 \tilde \UI_{jk}^{(c_m)} +
(\bar \UI_{jk}^{(c_m)})^2 \tilde \UI_{ik}^{(r_m)} +
\tilde \UI_{jk}^{(c_m)} \tilde \UI_{ik}^{(r_m)}
\right ) \right ], \notag
\end{align}
where the sum for $q_m$ is over all observed entries.

\paragraph{Updated for the pseudo-data:}

For each matrix with non-Gaussian data we update the pseudo-data
$\Z_m$ using
\begin{align}
\Xib_m &= E[\U_{r_m}] E[\U_{c_m}]^T, \label{eq:pseudodata} \\
\Z_m &= (\Xib_m - f_m'(\Xib_m)/\kappa_m), \notag
\end{align}
where the updates are element-wise and independent for each matrix.
Here $f_m'(\Xib_m)$ is the derivative of the $m$-th link function $-\log p(\X_m|\U_{r_m}\U_{c_m}^T)$
and $\kappa_m$ is the maximum value of the second derivative of the same
function.

\section*{MAP estimation}

These update rules can be easily modified to provide the MAP
estimate instead; the modifications mostly consist of dropping
the variance terms and the resulting updates are not repeated here.
Similarly, the updates are easy to modify for learning CMF models
without private factors, by coercing $\alpha_{ek}$ into $\alpha_k$.

\end{document}